\documentclass{article}

\usepackage{PRIMEarxiv}

\usepackage[utf8]{inputenc} 
\usepackage[T1]{fontenc}    
\usepackage{hyperref}       
\usepackage{url}            
\usepackage{booktabs}       
\usepackage{amsfonts}       
\usepackage{nicefrac}       
\usepackage{microtype}      
\usepackage{lipsum}
\usepackage{fancyhdr}       
\usepackage{graphicx}       
\graphicspath{{media/}}
\usepackage{enumitem}
\usepackage{amsmath, amssymb, amsthm}
\usepackage{caption}
\usepackage{tikz}
\usepackage{bm}
\usepackage{caption}
\usepackage{subcaption}
\usetikzlibrary{decorations.pathreplacing, calligraphy}

\title{Pairwise Markov Chains for Volatility Forecasting}

\author{
 Elie Azeraf, PhD \\
  Azeraf Financial Consulting\\
  Aix-en-Provence, France \\
  \texttt{elie.azeraf@azeraf-fc.com}
}

\begin{document}
\maketitle

\begin{abstract}
The Pairwise Markov Chain (PMC) is a probabilistic graphical model extending the well-known Hidden Markov Model. This model, although highly effective for many tasks, has been scarcely utilized for continuous value prediction. This is mainly due to the issue of modeling observations inherent in generative probabilistic models. In this paper, we introduce a new algorithm for prediction with the PMC. On the one hand, this algorithm allows circumventing the feature problem, thus fully exploiting the capabilities of the PMC. On the other hand, it enables the PMC to extend any predictive model by introducing hidden states, updated at each time step, and allowing the introduction of non-stationarity for any model. We apply the PMC with its new algorithm for volatility forecasting, which we compare to the highly popular GARCH(1,1) and feedforward neural models across numerous pairs. This is particularly relevant given the regime changes that we can observe in volatility. For each scenario, our algorithm enhances the performance of the extended model, demonstrating the value of our approach.
\end{abstract}


\section{Introduction}

\paragraph{}
Market volatility \cite{shiller1992market} is one of the most crucial concepts in finance. This notion, capturing stock fluctuations, is pivotal across various financial applications including options pricing, portfolio management, and risk management \cite{hull1993options}. For example, volatility is used to price options with the Black-Scholes \cite{black1973pricing} formulae, and it is primordial on the outcome obtained. Therefore, given its central role in the world of financial engineering, reliably predicting volatility is of paramount importance and can benefit all actors in the financial industry. 

\paragraph{}
There exist numerous ways to define it. Considering the stock 1-minute open prices $S_t$ and $T$ as the number of minutes, in this paper, we use the following formula corresponding to Historic Volatility:\cite{poon2003forecasting}:
\begin{align*}
    \sigma^2_t = \sqrt{\frac{1}{T} \sum\limits_{\tau = (t - 1) \times T}^{t \times T} u^2_{\tau}}.
\end{align*}
Here, $u_{\tau}$ represents the 1-minute log-return, defined as $u_{\tau} = \log\left(\frac{S_{\tau + 1}}{S_\tau}\right)$.

\paragraph{}
A well-known model for dealing with this concept is the GARCH(1, 1) \cite{bollerslev1986generalized}, introduced in 1986, which allows for its assessment, analysis, and prediction. Since then, a vast number of models have tackled this task, such as extensions of the GARCH model \cite{nelson1991conditional, engle1986modelling}, neural network-based models \cite{ge2022neural, bucci2020cholesky, vidal2020gold, yang2020html}, or those based on probabilistic graphical models like the Hidden Markov Chain (HMC) \cite{augustyniak2019new, luo2022forecasting}, also known as the Hidden Markov Model.

\paragraph{}
In this project, we introduce a novel approach to forecast volatility using the Pairwise Markov Chain (PMC) model \cite{pieczynski2003pairwise}. This model, a probabilistic graphical model expanding on the HMC, has rarely been employed for time series prediction \cite{escudier2024forecasting}. This limited usage is attributed to the difficulty in handling the features of observed variables \cite{ng2001discriminative, azeraf2021highly}, a problem that we will describe in detail in the next section, and for which we will propose a solution.

\paragraph{}
Our contribution lies in presenting a new algorithm tailored for time series prediction utilizing the PMC model, capable of handling observations' features without constraints. Unlike conventional prediction algorithms, our approach extends beyond mere forecasting by integrating hidden states into any prediction model. Specifically, the PMC-$f$ model enables dynamic adjustment of the parameters of a given forecasting model $f$ based on the observed time series and its underlying properties.

\paragraph{}
The main contributions of our paper are the following:
\begin{enumerate}[label=(\roman*), itemsep=0em, topsep=0pt, partopsep=0pt]
    \item We develop a new algorithm for PMC forecasting, allowing the use of PMC without feature constraints.
    \item We demonstrate how the PMC extends any forecasting model for time series prediction.
    \item We establish the empirical superiority of PMC for volatility forecasting using this new methodology.
\end{enumerate}

\paragraph{}
This article is organized as follows: First, the next section provides a literature review, presenting the main models for volatility forecasting and detailing the Pairwise Markov Chain. Section 3 is devoted to our main contribution, presenting the new prediction algorithm using PMC, followed by Section 4, which presents empirical results on volatility for different pairs. Finally, our paper concludes with a discussion and conclusion.

\section{Theoretical Framework}

\subsection{Main models for volatility forecasting}

\paragraph{}
The most popular model for processing and predicting volatility is the GARCH(1, 1) model \cite{bollerslev1986generalized}. Given the log-return at time $t$, $u_t$, and the volatility at time $t$, $\sigma^2_t$, it models the next volatility $\sigma^2_{t + 1}$ as follows:
\begin{equation}
    \sigma^2_{t + 1} = \omega + \alpha u_t^2 + \beta \sigma^2_t + \epsilon_{t + 1}
\end{equation}
with $\epsilon_{t + 1}$ being a random variable with zero mean, and $(\alpha, \beta, \omega)$ representing the model's parameters of interest.

\paragraph{}
Therefore, forecasting with GARCH(1, 1) involves computing
\begin{equation}
    \mathbb{E}[\sigma^2_{t + 1}] = \omega + \alpha u_t^2 + \beta \sigma^2_t
\end{equation}
It can be viewed as a linear combination of the input data to produce a unique output.

This model has been widely extended to incorporate additional features, such as the EGARCH \cite{nelson1991conditional}, GARCH-M \cite{engle1986modelling}, and several other models, enhancing its predictive capabilities \cite{lundbergh2002evaluating, terasvirta2009introduction, bauwens2006multivariate, francq2019garch, franses1996forecasting, chen2010forecasting, klaassen2002improving}.

\paragraph{}
Another widely used family of models for this task is neural network-based ones \cite{lecun2015deep, goodfellow2016deep}, with studies covering various neural architectures \cite{ge2022neural, bucci2020cholesky, vidal2020gold, yang2020html}. For our work, we focus on the Feedforward Neural Network (FNN), which comprises a succession of linear combinations followed by non-linear functions, called activation functions. The input data are referred to as the input layer, the returned data as the output layer, and the intermediate variables resulting from linear combinations and activation functions as the hidden layers.

\paragraph{}
Given $t$, for our actual work, we consider the volatility $\sigma_t^2$ and the squared log-return $u_t^2$ as the input layer, and we use three different FNNs:
\begin{itemize}[itemsep=0em, topsep=0pt, partopsep=0pt]
    \item FNN(2): composed of the input layer of dimension 2 for our case, a hidden layer of dimension 2, and the output layer, which is of dimension 1 in our case.
    \item FNN(3): composed of the input layer, a hidden layer of dimension 3, and the output layer.
    \item FNN(2, 3): composed of the input layer, two hidden layers of dimension 3, and the output layer.
\end{itemize}
We use the hyperbolic tangent as activation functions. As we aim to predict a continuous random variable, our FNNs do not conclude with an activation function in the output layer.

\begin{figure}[t]
\centering
\begin{tikzpicture}
[font=\small, inner sep=0pt, visible/.style = {circle,draw = blue!15, fill = blue!10, thick,minimum size = 1cm, rounded corners}, hidden/.style = {circle,draw=black!35,fill=black!30,thick,minimum size=1cm, rounded corners}, scale = 1]
\node at (-4,0) (y1) [visible]  {$Y_1$};
\node at (-2,0) (y2) [visible]  {$Y_2$};
\node at (0,0) (y3) [visible]  {$Y_3$};
\node at (2,0) (y4) [visible]  {$Y_4$};

\draw [->, >=stealth] (y1) to (y2);
\draw [->, >=stealth] (y2) to (y3);
\draw [->, >=stealth] (y3) to (y4);
\end{tikzpicture}
\caption{Oriented Probabilistic Graph of the Markov Chain.}
\label{fig_mc}
\end{figure}
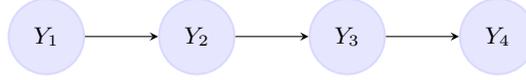

\paragraph{}
All the models presented in this subsection model $\sigma^2_{t + 1}$ as a function of $\sigma^2_{t}$ and $u_t^2$, and assume that:
\begin{align}
    \sigma^2_{t + 1} = f(\sigma^2_{t}, u_t^2) + \epsilon_t,
    \label{eq_f_only}
\end{align}
where $f$ is a model and $\epsilon_t$ is a random variable with mean equal to $0$ and independent from the other random variables. We consider the random variable $Y_t$ taking its values in $\mathbb{R}^{2}$ such that the realization $y_t$ of $Y_t$ is $(\sigma^2_{t + 1}, u_t^2)$. With each model presented in this subsection, the observed process $Y_{1:T} = (Y_1, ..., Y_T)$ forms a Markov chain, and its probabilistic graph is represented in figure \ref{fig_mc}.

\subsection{The Pairwise Markov Chain model}

\subsubsection{Definition}

\begin{figure}[t]
\centering
\begin{tikzpicture}
[font=\small, inner sep=0pt, visible/.style = {circle,draw = blue!15, fill = blue!10, thick,minimum size = 1cm, rounded corners}, hidden/.style = {circle,draw=black!35,fill=black!30,thick,minimum size=1cm, rounded corners}, scale = 1]
\node at (-4,-2) (x1) [hidden] {$X_1$};
\node at (-4,0) (y1) [visible]  {$Y_1$};
\draw [->, >=stealth] (x1) to (y1);
                    
\node at (-2,-2) (x2) [hidden] {$X_2$};
\node at (-2,0) (y2) [visible]  {$Y_2$};
\draw [->, >=stealth] (x2) to (y2);
                    
\node at (0,-2) (x3) [hidden] {$X_3$};
\node at (0,0) (y3) [visible]  {$Y_3$};
\draw [->, >=stealth] (x3) to (y3);
                    
\node at (2, -2) (x4) [hidden] {$X_4$};
\node at (2,0) (y4) [visible]  {$Y_4$};
\draw [->, >=stealth] (x4) to (y4);
                    
\draw [->, >=stealth] (x1) to (x2);
\draw [->, >=stealth] (x2) to (x3);
\draw [->, >=stealth] (x3) to (x4);

\draw [->, >=stealth] (y1) to (y2);
\draw [->, >=stealth] (y2) to (y3);
\draw [->, >=stealth] (y3) to (y4);

\draw [->, >=stealth] (x1) to (y2);
\draw [->, >=stealth] (x2) to (y3);
\draw [->, >=stealth] (x3) to (y4);

\draw [->, >=stealth] (y1) to (x2);
\draw [->, >=stealth] (y2) to (x3);
\draw [->, >=stealth] (y3) to (x4);
\end{tikzpicture}
\caption{Oriented Probabilistic Graph of the Pairwise Markov Chain.}
\label{fig_pmc}
\end{figure}
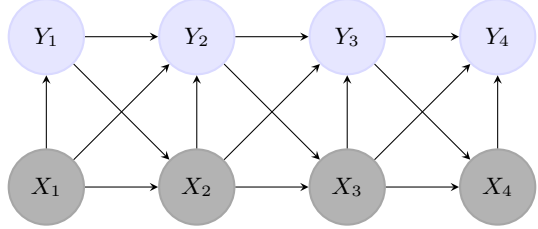

\paragraph{}
The Pairwise Markov Chain (PMC) \cite{pieczynski2003pairwise} is a probabilistic graphical model that extends the well-known HMC \cite{stratonovich1965conditional, baum1966statistical, cappe2006inference, rabiner1986introduction}, improving its performance in many fields \cite{azeraf2021highly, derrode2004signal, gorynin2017pairwise, gorynin2016optimal}. It considers the hidden random sequence $X_{1:T} = (X_1, ..., X_T)$, where $X_t$ takes its value in $\Lambda_X$, and the observed sequence $Y_{1:T} = (Y_1, ..., Y_T)$, where $Y_t$ takes its value in $\Omega_Y$. For this work, we consider $\Lambda_X = (\lambda_1, ..., \lambda_N)$ as a finite discrete set, and $\Omega_Y = \mathbb{R}^d$, where $d \geq 1$. We also consider a homogeneous PMC, meaning that the laws defining it do not depend on $t$. The PMC defines the distribution of $(X_{1:T}, Y_{1:T})$ as follows:
\begin{equation}
    p(x_{1:T}, y_{1:T}) = p(x_1, y_1) \prod_{t = 1}^{T - 1} p(x_{t + 1}, y_{t + 1} | x_t, y_t).
\end{equation}
Its probabilistic graph is depicted in figure \ref{fig_pmc}. As one can observe, it extends the model shown in figure \ref{fig_mc} by adding a hidden process.

\begin{figure}[t]
\centering
\begin{tikzpicture}
[font=\small, inner sep=0pt, visible/.style = {circle,draw = blue!15, fill = blue!10, thick,minimum size = 1cm, rounded corners}, hidden/.style = {circle,draw=black!35,fill=black!30,thick,minimum size=1cm, rounded corners}, scale = 1]
\node at (-4,-2) (x1) [hidden] {$X_1$};
\node at (-4,0) (y1) [visible]  {$Y_1$};
\draw [->, >=stealth] (x1) to (y1);
                    
\node at (-2,-2) (x2) [hidden] {$X_2$};
\node at (-2,0) (y2) [visible]  {$Y_2$};
\draw [->, >=stealth] (x2) to (y2);
                    
\node at (0,-2) (x3) [hidden] {$X_3$};
\node at (0,0) (y3) [visible]  {$Y_3$};
\draw [->, >=stealth] (x3) to (y3);
                    
\node at (2, -2) (x4) [hidden] {$X_4$};
\node at (2,0) (y4) [visible]  {$Y_4$};
\draw [->, >=stealth] (x4) to (y4);
                    
\draw [->, >=stealth] (x1) to (x2);
\draw [->, >=stealth] (x2) to (x3);
\draw [->, >=stealth] (x3) to (x4);
\end{tikzpicture}
\caption{Oriented Probabilistic Graph of the Hidden Markov Chain.}
\label{fig_hmc}
\end{figure}
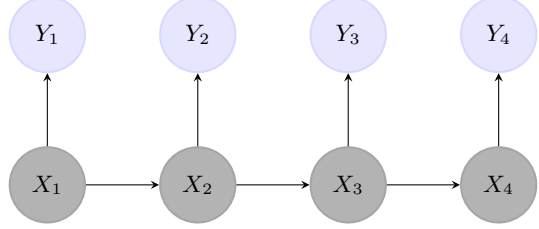

\paragraph{Remark} Considering the same random variables, the probabilistic law of the HMC is:
\begin{equation}
    p(x_{1:T}, y_{1:T}) = p(x_1) \prod_{t = 1}^{T - 1} p(x_{t + 1} | x_t) \prod_{t = 1}^{T} p(y_{t} | x_t),
\end{equation}
and its graph is in figure \ref{fig_hmc}. Therefore, we see that the PMC is a HMC if and only if $p(x_{t + 1}, y_{t + 1} | x_t, y_t) = p(x_{t + 1} | x_t) p(y_{t + 1} | x_{t + 1})$.

\subsubsection{Usual Forecasting Algorithm}
\label{usual_forecasting_algorithm}

\paragraph{}
Given $y_{1:t}$, forecasting with the PMC model involves computing $\mathbb{E}[y_{t + 1} | y_{1:t}]$. This value is typically computed as follows:
\begin{align*}
    \mathbb{E}[y_{t + 1} | y_{1:t}] &= \mathbb{E}\left[ \mathbb{E}\left[ x_t, x_{t + 1}, y_{t + 1} | y_{1:t} \right] \right] \\
    &= \sum_{x_t} \sum_{x_{t + 1}} p(x_t, x_{t + 1} | y_{1:t}) \mathbb{E}[y_{t + 1} | x_t, y_t, x_{t + 1}] \\
    &= \sum_{x_t} \sum_{x_{t + 1}} p(x_t | y_{1:t}) p(x_{t + 1} | x_t, y_{t}) \mathbb{E}[y_{t + 1} | x_t, y_t, x_{t + 1}].
\end{align*}
$p(x_t | y_{1:t})$ is computed using the Bayes' law:
\begin{equation*}
    p(x_t | y_{1:t}) = \frac{p(x_t, y_{1:t})}{\sum\limits_{x'_t} p(x'_t, y_{1:t})},
\end{equation*}
computing $p(x_t, y_{1:t})$ recursively:
\begin{itemize}[itemsep=0em, topsep=0pt, partopsep=0pt]
    \item For $t = 1$:
    \begin{align*}
    p(x_1, y_1) &= p(x_1) p(y_1 | x_1)
    \end{align*}

    \item For $t > 1$:
    \begin{align*}
    p(x_{t + 1}, y_{1:t +1}) &= \sum\limits_{x_t} p(x_t, y_{1:t}, x_{t + 1}, y_{t + 1}) \\
    &= \sum\limits_{x_t} p(x_t, y_{1:t}) p(x_{t + 1} | x_t, y_t) p(y_{t + 1} | x_t, y_t, x_{t + 1}).
\end{align*}
\end{itemize}

\paragraph{}
Therefore, to apply this algorithm, one needs, for all $t$, $p(x_t)$, $p(y_t | x_t)$, $p(x_{t + 1} | x_t, y_t)$, and $p(y_{t + 1} | x_t, y_t, x_{t + 1})$. Since we do not observe the hidden sequence $x_{1:T}$ in the training set, the Expectation-Maximization algorithm \cite{dempster1977maximum} can be used to learn these parameters.

\subsubsection{Handling observations' features}

\paragraph{}
Using the above algorithm, one encounters a well-established problem inherent in algorithms that need to learn the law of observations, as may be the case with probabilistic generative models. Indeed, we have to learn two observation laws: $p(y_t | x_t)$ and $p(y_{t + 1} | x_t, y_t, x_{t + 1})$.

\paragraph{}
First, we must assume the law. Although a Gaussian distribution is commonly assumed for its generality and computational simplicity, the actual nature of the distribution may differ significantly

\paragraph{}
Moreover, the number of parameters is also a problem. If we assume $p(y_t | x_t)$ follows a Gaussian distribution, we have to learn all the means, implying $d \times N$ parameters, and the variance matrices, with $\frac{d (d + 1)}{2} \times N$ parameters, resulting in a quadratic dependence. For large values of $d$, this implies a very large number of parameters. This leads to computational challenges and increased demand for data, potentially imposing limitations on the feasibility of the algorithm.

\paragraph{}
Conversely, when focusing on learning laws such as $p(x_t | y_t)$, encountered in discriminative models, and assuming a logistic regression model, the parameterization simplifies significantly. In this case, only $d + 1$ parameters need to be learned. This stark reduction in the number of parameters not only streamlines the learning process but also minimizes the risk of overfitting, particularly in situations where the dimensionality of the data is high. The literature largely describes this problem and compares the benefits of using the latest algorithms \cite{ng2001discriminative, lafferty2001conditional, mccallum2000maximum}.

\paragraph{}
Whereas it was accepted that this problem was encountered with generative models, and discriminative ones should be preferred, recent works show that this is not the case and prove that any generative model can define a classifier without using any observation laws \cite{azeraf2022deriving}. For example, \cite{azeraf2020hidden, azeraf2023equivalence} present the Discriminative Forward-Backward algorithm, allowing the computation of $p(x_t | y_{1:T})$ with the HMC without using observation laws, and significantly improving the results. The same work is done for the Naive Bayes model in \cite{azeraf2021using}.

\section{Methodology}

\subsection{New algorithm}
\label{new_algorithm}

In \ref{usual_forecasting_algorithm}, we have seen that $\mathbb{E}[y_{t + 1} | y_{1:t}]$ can be computed using the PMC model as:
\begin{align}
    \mathbb{E}[y_{t + 1} | y_{1:t}] &= \sum_{x_t} p(x_t | y_{1:t}) \sum_{x_{t + 1}} p(x_{t + 1} | x_t, y_{t}) \mathbb{E}[y_{t + 1} | x_t, y_t, x_{t + 1}].
    \label{expectation_pmc}
\end{align}

\paragraph{Proposition 3.1} Our new algorithm consists of computing $p(x_t | y_{1:t})$ in a different way, without relying on any observation laws.
\begin{itemize}[itemsep=0em, topsep=0pt, partopsep=0pt]
    \item For $t = 1$:
    \begin{align*}
        p(x_1 | y_1) \text{ is given};
    \end{align*}

    \item For $t \geq 1$:
    \begin{align}
        p(x_{t + 1} | y_{1:t + 1}) = \frac{\gamma_{t + 1}(x_{t + 1})}{\sum\limits_{x'_{t + 1}} \gamma_{t + 1}(x'_{t + 1})}
    \end{align}
    with
    \begin{align}
        \gamma_{t + 1}(x_{t + 1}) = \sum\limits_{x_t} p(x_t | y_{1:t}) \frac{p(x_t | y_t, y_{t + 1})}{p(x_t | y_t)} p(x_{t + 1} | x_t, y_t, y_{t + 1}).
        \label{alpha_recurence}
    \end{align}
\end{itemize}
Thus, computing $\mathbb{E}[y_{t + 1} | y_{1:t}]$ never relies on observation laws, allowing PMC to be used for forecasting while avoiding the feature problem.

\paragraph{Proof:} First, we express the probability $p(x_{t + 1}, y_{1:t + 1})$ as follows:
\begin{align*}
    p(x_{t + 1}, y_{1:t + 1}) &= \sum_{x_t} p(x_t, x_{t + 1}, y_{1:t}, y_{t + 1}) \\
    &= p(y_{1:t}) \sum_{x_t} p(x_t | y_{1:t}) p(y_{t + 1} | x_t, y_t) p(x_{t + 1} | x_t, y_t, y_{t + 1}) \\
    &= p(y_{1:t}) \sum_{x_t} p(x_t | y_{1:t}) \frac{p(x_t, y_t, y_{t + 1})}{p(x_t, y_t)} p(x_{t + 1} | x_t, y_t, y_{t + 1}) \\
    &= p(y_{1:t}) p(y_{t + 1} | y_t) \sum_{x_t} p(x_t | y_{1:t}) \frac{p(x_t | y_t, y_{t + 1})}{p(x_t | y_t)} p(x_{t + 1} | x_t, y_t, y_{t + 1}) \\
    &= p(y_{1:t}) p(y_{t + 1} | y_t) \gamma_{t + 1}(x_{t + 1}).
\end{align*}

Therefore, 
\begin{align*}
    p(x_{t + 1} | y_{1:t + 1}) &= \frac{p(x_{t + 1}, y_{1:t + 1})}{\sum\limits_{x'_{t + 1}} p(x'_{t + 1}, y_{1:t + 1})} 
    = \frac{\gamma_{t + 1}(x_{t + 1})}{\sum\limits_{x'_{t + 1}} \gamma_{t + 1}(x'_{t + 1})},
\end{align*}
ending the proof.

\paragraph{Remark 3.1} From the Proposition below, we can deduce the same algorithm for the HMC:
\begin{align*}
    \mathbb{E}[y_{t + 1} | y_{1:t}] &= \sum_{x_t} p(x_t | y_{1:t}) \sum_{x_{t + 1}} p(x_{t + 1} | x_t) \mathbb{E}[y_{t + 1} | x_{t + 1}],
\end{align*}
with
\begin{align*}
    p(x_1 | y_1) & \text{ is given}; \\
    p(x_{t + 1} | y_{1:t + 1}) &= \frac{\delta_{t + 1}(x_{t + 1})}{\sum\limits_{x'_{t + 1}} \delta_{t + 1}(x'_{t + 1})} \\
    \text{with } \delta_{t + 1}(x_{t + 1}) &= \sum\limits_{x_t} p(x_t | y_{1:t}) \frac{p(x_{t + 1} | x_t)}{p(x_{t + 1})} p(x_{t + 1} | y_{t + 1}).
\end{align*}
The proof may be directly deducted from the Proof below.

\subsection{How the PMC extends other models}

\begin{figure}
  \begin{subfigure}{.45\textwidth}
    \centering
    \begin{tikzpicture}
      \node[draw, circle, inner sep=6pt] (yt) at (0,0) {$y_t$};
      \node[draw, rectangle, inner sep=5pt, minimum width=1cm] (f) at (0,2) {$f$};
      \node[draw, circle] (yt_plus_1) at (0,4) {$\hat{y}_{t + 1}$};

      \draw[->, >=stealth] (yt) -- (f);
      \draw[->, >=stealth] (f) -- (yt_plus_1);

      \draw[dotted, line width=1pt] (-1,1) rectangle (1,3);
    \end{tikzpicture}
    \caption{With the model $f$}
    \label{computing_process_f}
  \end{subfigure}
  \begin{subfigure}{.45\textwidth}
    \centering
    \begin{tikzpicture}
      \node[draw, circle] (yt) at (0,0) {$y_t$};
      \node[draw, circle] (yprev) at (-2,0) {$y_{1:t-1}$};
      \node[draw, rectangle, inner sep=5pt, minimum width=1cm] (f) at (0.75,2) {$f_{x_t}$};
      \node[draw, rectangle, inner sep=6pt] (alpha) at (-1.25,2) {$p(x_t | y_{1:t})$};
      \node[draw, circle] (times) at (0,3) {$\times$};
      \node[draw, circle] (sum) at (0,4.5) {$\sum$};
      \node[draw, circle] (yt_plus_1) at (0,6) {$\hat{y}_{t + 1}$};
      \draw[->, >=stealth] (yt) -- (f);
      \draw[->, >=stealth] (yprev) -- (alpha);
      \draw[->, >=stealth] (f) -- (times);
      \draw[->, >=stealth] (yt) -- (alpha);
      \draw[->, >=stealth] (f) -- (times);
      \draw[->, >=stealth] (alpha) -- (times);
      \draw[->, >=stealth] (times) -- (sum);
      \draw[->, >=stealth] (sum) -- (yt_plus_1);
      \node at (2.5,2.25) {$\times N$};
      \draw[dotted, line width=1pt] (-3,1) rectangle (2,3.5);
    \end{tikzpicture}
    \caption{With the model PMC(N)-$f$}
    \label{computing_process_pmcf}
  \end{subfigure}
  \captionof{figure}{Computing the prediction $\hat{y}_{t + 1}$ with the models $f$ and PMC(N)-$f$.}
\end{figure}
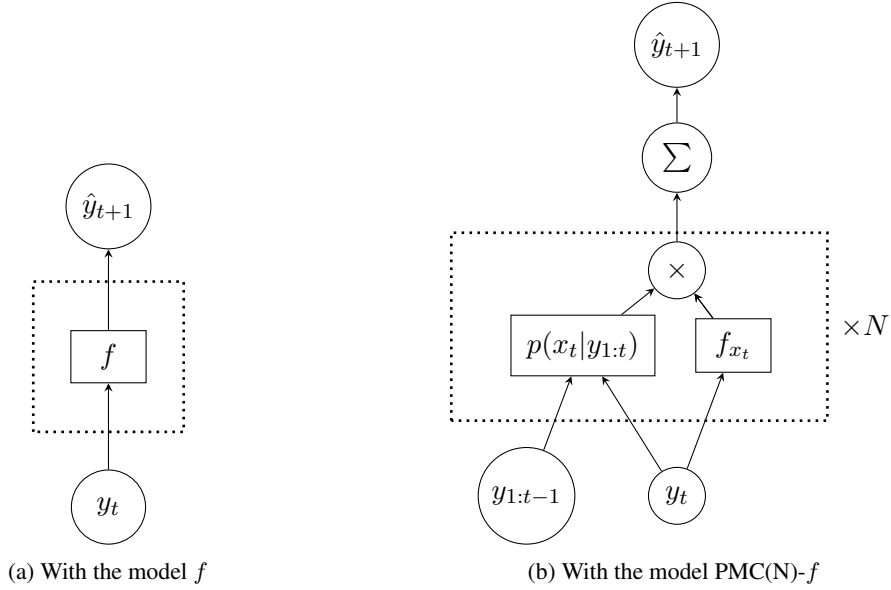

\paragraph{}
Given a forecasting model such as the GARCH(1, 1), the FNN, or any other, we define $f: \mathbb{R}^{d} \longrightarrow \mathbb{R}^{d'}$ as its function. Therefore, given the observations $y_{1:t}$, we estimate the next observed variable as $\hat{y}_{t + 1} = f(y_{t})$. The computing process is illustrated in figure \ref{computing_process_f}.

\paragraph{}
With the PMC and its forecasting algorithm described in \ref{new_algorithm}, one can write formula \ref{expectation_pmc} as follows without loss of generality:
\begin{align*}
    \mathbb{E}[y_{t + 1} | y_{1:t}] &= \sum_{x_t} p(x_t | y_{1:t}) \sum_{x_{t + 1}} p(x_{t + 1} | x_t, y_{t}) \mathbb{E}[y_{t + 1} | x_t, y_t, x_{t + 1}]. \\
    &= \sum_{x_t} p(x_t | y_{1:t}) g(x_t, y_t)
\end{align*}
with $g: \Lambda_X \times \Omega_Y \longrightarrow \Omega_Y$. Moreover, one may set $g(x_t, y_t) = f_{x_t}(y_t)$, where $f_{x_t}$ is a function similar to $f$ but with different parameter values. Therefore, with the PMC model having $N$ hidden states and our new algorithm, given $N$ functions $f_i$ similar to the preceding model $f$, the estimation of the next observed variable is:
\begin{align}
    \hat{y}_{t + 1} = \sum_{x_t} p(x_t | y_{1:t}) f_{x_t}(y_{t}).
    \label{pmc_f}
\end{align}

\paragraph{}
Therefore, prediction with the PMC can be seen as the weighted average of $N$ models $f$, where the weights $p(x_t | y_{1:t})$ are calculated based on the entire observed sequence and are computed sequentially with complex dependencies between all the random variables. This computing process is illustrated in figure \ref{computing_process_pmcf}.

\paragraph{}
This method offers several benefits:
\begin{itemize}[itemsep=0em, topsep=0pt, partopsep=0pt]
    \item The weights at time $t + 1$ differ from those at time $t$, thanks to the information provided by the new observation at time $t$;
    \item The computation of the weights at time $t + 1$ is fast, depending only on the weights and the observation at time $t$, thanks to the Markov chain structure;
    \item This approach generalizes the use of only the model $f$, as having only one hidden state ($N = 1$) is sufficient to recover the latter. Therefore, we retain the benefits of the initial model, such as the mean-reversion behavior of the GARCH(1, 1), for example.
\end{itemize}

\paragraph{Remark 3.2.} The PMC with our new algorithm and the Regime Switching Models \cite{klaassen2002improving, hamilton2010regime} may seem similar, both introducing the notion of hidden states to incorporate non-stationarity. However, our approach is much more general. The Regime Switching Model relies on modeling with far fewer dependencies between random variables, similar to the HMC, sometimes adding a dependency linking the observed random variables. In contrast, the PMC is more general, with more dependencies between the random variables.

\section{Results}

\paragraph{}
In this section, we will apply our new PMC algorithm to empirically observe its benefits, depending on base models. For example, we will assess the performance of the GARCH(1, 1) model for volatility forecasting, and then we will combine this model with a PMC having $N$ hidden states, as described in \ref{pmc_f}, to study the advantages of being able to switch between different states.

\paragraph{}
In the context of our empirical study, we will use the GARCH(1, 1), the FNN(2), the FNN(3), and the FNN(2, 3) as base models. These models will be combined with PMCs having $2$, $3$, and $4$ hidden states. The model associating the PMC with $N$ hidden states and a model $f$ is denoted PMC($N$)-$f$.

\subsection{Data}

\begin{figure}[t]
    \centering

    \begin{subfigure}[b]{0.24\textwidth}
        \centering
        \includegraphics[scale=0.25]{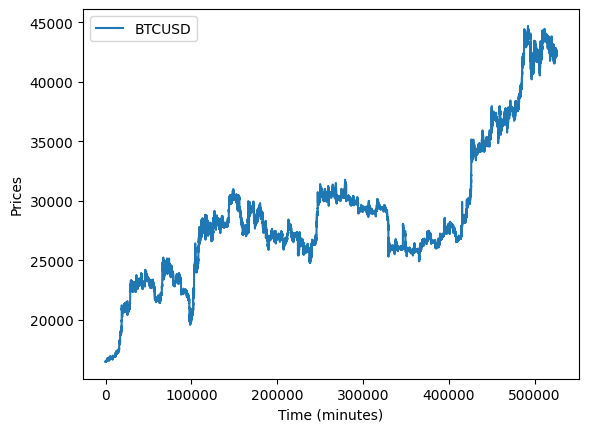}
        \caption{BTC}
    \end{subfigure}
    \begin{subfigure}[b]{0.24\textwidth}
        \centering
        \includegraphics[scale=0.25]{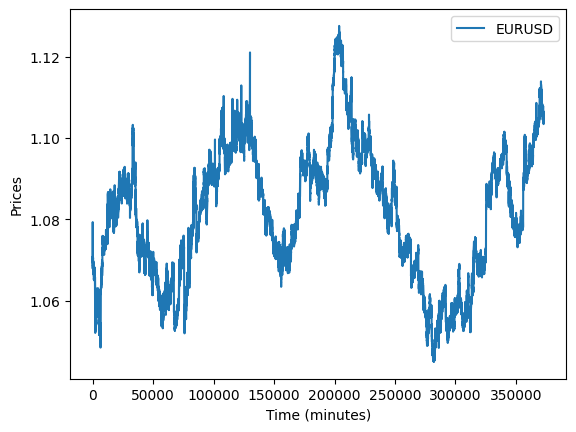}
        \caption{EUR/USD}
    \end{subfigure}
    \begin{subfigure}[b]{0.24\textwidth}
        \centering
        \includegraphics[scale=0.25]{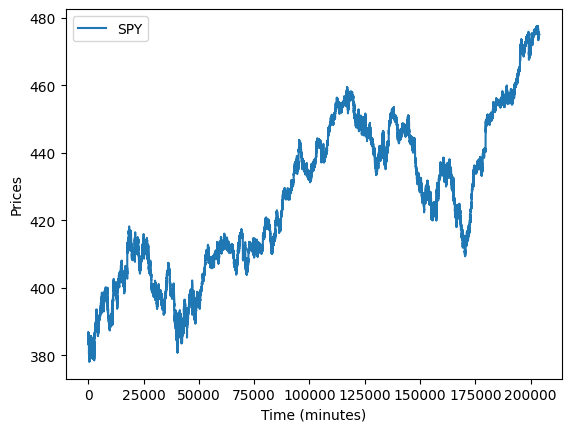}
        \caption{SPY}
    \end{subfigure}
    \begin{subfigure}[b]{0.24\textwidth}
        \centering
        \includegraphics[scale=0.25]{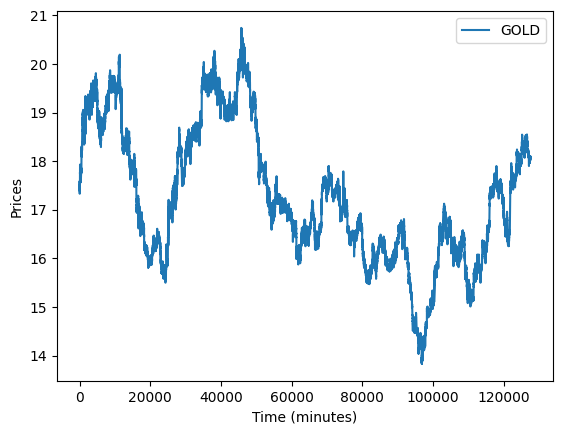}
        \caption{GOLD}
    \end{subfigure}

    \caption{Our four pairs one-minute open-data during 2023.}
    \label{open_data_four_pairs}
\end{figure}

\paragraph{}
We focus our empirical study on four different pairs: BTC/USD, EUR/USD, SPY, and GOLD\footnote{Data may be freely downloaded at \href{https://polygon.io/}{https://polygon.io/}.}. We will apply our algorithms to data throughout the year 2023. The open prices of each pair, with 1-minute intervals, are represented in figure \ref{open_data_four_pairs}.

\begin{figure}[t]
    \centering

    \begin{subfigure}[b]{0.24\textwidth}
        \centering
        \includegraphics[scale=0.25]{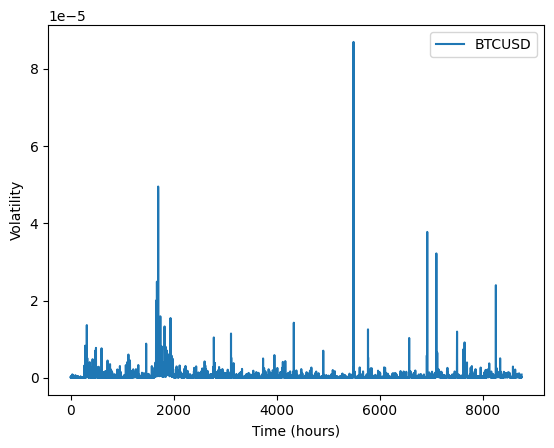}
        \caption{BTC}
    \end{subfigure}
    \begin{subfigure}[b]{0.24\textwidth}
        \centering
        \includegraphics[scale=0.25]{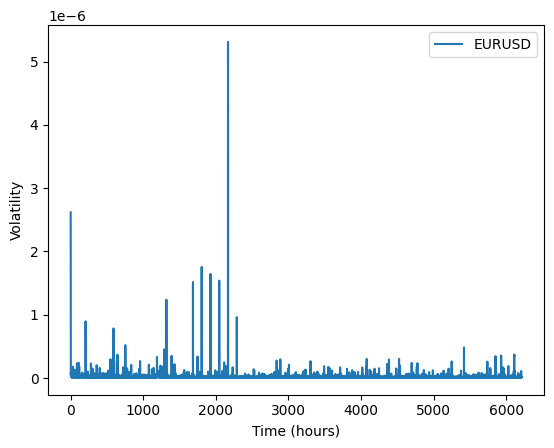}
        \caption{EUR/USD}
    \end{subfigure}
    \begin{subfigure}[b]{0.24\textwidth}
        \centering
        \includegraphics[scale=0.25]{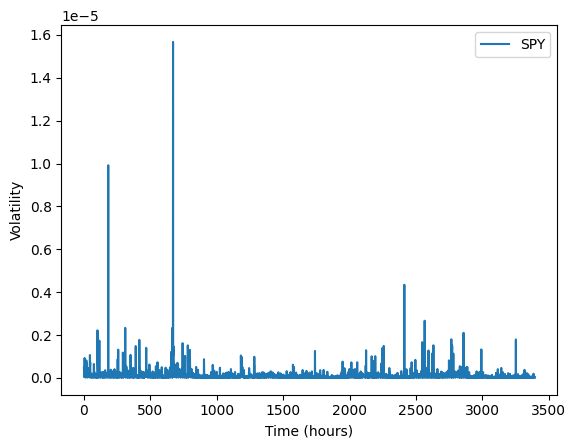}
        \caption{SPY}
    \end{subfigure}
    \begin{subfigure}[b]{0.24\textwidth}
        \centering
        \includegraphics[scale=0.25]{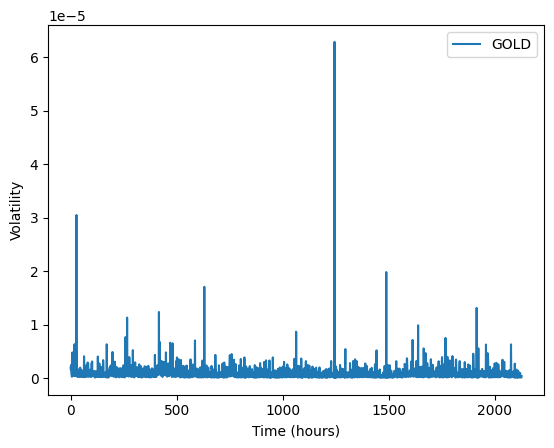}
        \caption{GOLD}
    \end{subfigure}

    \caption{Our four pairs volatility (calculated over each 60-minute interval) during 2023.}
    \label{volatility_four_pairs}
\end{figure}

\paragraph{}
Our variable of interest, the volatility, is computed over 60-minute intervals. To recall, given the open prices $S_{1:T} = (S_1, ..., S_T)$, we compute the 1-minute log-returns $u_{1:T} = (u_1, ..., u_{T - 1})$ such that $u_t = \log\left(\frac{S_{t + 1}}{S_t}\right)$. Then, we compute volatility data of the stock as follows, for each $t$:
\begin{align*}
    \sigma^2_t = \sqrt{\frac{1}{60} \sum\limits_{\tau = 60 (t - 1)}^{60t} u_{\tau}^2}.
\end{align*}
The volatility data of our four pairs are represented in figure \ref{volatility_four_pairs}. 

\paragraph{}
For each $t$, our goal is to predict the next volatility $\sigma^2_{t + 1}$ given the observed volatility sequence $\sigma_{1:t}^2 = (\sigma_{1}^2, ..., \sigma_{t}^2)$ and the observed 60-minute log-returns sequence $u^{(60)}_{1:t} = (u^{(60)}_1, ..., u^{(60)}_{t})$, which are squared and computed such that $u^{(60)}_t = \log\left(\frac{S_{60 (t + 1)}}{S_{60 t}}\right)$. For computational purposes, the input and target data are centered and scaled, meaning that their mean is equal to 0 and their variance is equal to 1. Then, they are passed through the logarithm function to reduce discrepancies with outliers. These data are designated as "normalized data".

\paragraph{}
We evaluate a model's performance using the mean squared error (MSE), which is computed between the ground truth volatility data $\sigma_{1:T}^2$ and the estimated ones $\hat{\sigma}^2_{1:T}$:
\begin{align*}
    MSE(\sigma_{1:T}^2, \hat{\sigma}^2_{1:T}) = \frac{1}{T} \sum_{t = 1}^T \left(\sigma_{t}^2 - \hat{\sigma}^2_{t}\right)^2.
\end{align*}

\subsection{Training}

\paragraph{}
All the models are trained using the gradient descent algorithm with back-propagation \cite{lecun1988theoretical, lecun1989backpropagation}. We utilize the Adam optimizer \cite{kingma2014adam} and the MSE loss function, with a learning rate set to $0.05$. In terms of implementation details, all the code is written in Python using the PyTorch \cite{paszke2019pytorch} library, and the experiments are conducted using a 16GB CPU\footnote{All the code and experiments are available at \href{https://github.com/ElieAzeraf/Volatility_Prediction_PMC/tree/develop}{https://github.com/ElieAzeraf/Volatility\_Prediction\_PMC/tree/develop}}.

\paragraph{}
Regarding the PMC-based models, we utilize formula \ref{pmc_f} to forecast the next volatility, employing our new algorithm described in \ref{new_algorithm} to compute $p(x_t | y_{1:t})$. Without loss of generality, in \ref{alpha_recurence}, we model $\frac{p(x_t | y_t, y_{t + 1})}{p(x_t | y_t)} p(x_{t + 1} | x_t, y_t, y_{t + 1})$ with a neural network function, taking $x_t$, $y_t$, $x_{t + 1}$, and $y_{t + 1}$ as inputs and producing an output in $\mathbb{R}^+$. This simplifies the computation during back-propagation without loss of generality, as it is always possible to define the different probabilities given this function.

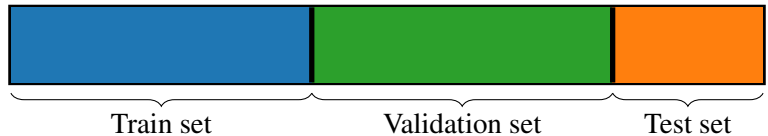
\begin{figure}
    \centering
    \begin{tikzpicture}
    \def\smallwidth{10}
    \def\smallheight{1}
    \def\framethickness{2pt}

    \definecolor{b}{HTML}{1f77b4} 
    \definecolor{g}{HTML}{2ca02c} 
    \definecolor{o}{HTML}{ff7f0e} 
    
    \draw[black, line width=\framethickness] (0,0) rectangle (\smallwidth,\smallheight);
    
    \fill[b] (0,0) rectangle ({0.4*\smallwidth},\smallheight);
    \fill[g] ({0.4*\smallwidth},0) rectangle ({0.8*\smallwidth},\smallheight);
    \fill[o] ({0.8*\smallwidth},0) rectangle (\smallwidth,\smallheight);
    
    \draw[black, line width=2pt] ({0.4*\smallwidth},0) -- ({0.4*\smallwidth},\smallheight);
    \draw[black, line width=2pt] ({0.8*\smallwidth},0) -- ({0.8*\smallwidth},\smallheight);
    
    \draw[decorate,decoration={brace,mirror,amplitude=5pt},yshift=-4pt] (0,0) -- ({0.4*\smallwidth},0) node [black,midway,below=4pt] {Train set};
    \draw[decorate,decoration={brace,mirror,amplitude=5pt},yshift=-4pt] ({0.4*\smallwidth},0) -- ({0.8*\smallwidth},0) node [black,midway,below=4pt] {Validation set};
    \draw[decorate,decoration={brace,mirror,amplitude=5pt},yshift=-4pt] ({0.8*\smallwidth},0) -- ({\smallwidth},0) node [black,midway,below=4pt] {Test set};
\end{tikzpicture}
\captionof{figure}{Illustration of the decomposition of the data.}
\label{splits_fig}
\end{figure}

\paragraph{}
A set of parameters is learned for each time series, which is divided into three sets:
\begin{itemize}[itemsep=0em, topsep=0pt, partopsep=0pt]
    \item The training set, comprising the first 40\% of the data;
    \item The validation set, comprising 40\% of the data in the series, extending up to 80\% of it;
    \item The test set, comprising the remaining data, ranging from 80\% of the data to the end of the series.
\end{itemize}
This decomposition is illustrated in figure \ref{splits_fig}. Data are not shuffled as we are dealing with temporal data.

\subsection{Results}

\paragraph{}
The results of all the models on normalized data are presented in Table \ref{results_normalized_data}, and on the original data in Table \ref{results_data}\footnote{For BTC, results have to be multiplied by $10^{-12}$, by $10^{-14}$ for SPY, by $10^{-12}$ for GOLD, and by $10^{-15}$ for EURUSD.}. All the experiments are reproduced five times, and we report the mean and the Gaussian 95\% confidence interval of the MSE. For each model, the best performance, which is the lowest, is highlighted in bold.

\begin{table}[t]
\centering
\begin{tabular}{c|c|c|c|c}
 {} & BTC & SPY & GOLD & EURUSD \\
 \hline
 \hline
 GARCH(1, 1) & $0.337 \pm 0.004$ & $0.807 \pm 0.014$ & $0.775 \pm 0.018$ & $0.625 \pm 0.000$\\
 PMC(2) - GARCH(1, 1) & $\bm{0.320} \pm 0.003$ & $0.780 \pm 0.004$ & $0.655 \pm 0.004$ & $\bm{0.609} \pm 0.005$ \\
 PMC(3) - GARCH(1, 1) & $0.324 \pm 0.005$ & $0.791 \pm 0.014$ & $\bm{0.636} \pm 0.013$ & $0.618 \pm 0.007$ \\
 PMC(4) - GARCH(1, 1) & $0.321 \pm 0.002$ & $\bm{0.770} \pm 0.011$ & $0.644 \pm 0.014$ & $0.616 \pm 0.005$ \\
 \hline
 FNN(2) & $0.331 \pm 0.002$ & $0.795 \pm 0.004$ & $0.744 \pm 0.20$ & $0.620 \pm 0.001$ \\
 PMC(2) - FNN(2) & $0.321 \pm 0.001$ & $\bm{0.779} \pm 0.005$ & $0.660 \pm 0.19$ & $0.620 \pm 0.005$\\ 
 PMC(3) - FNN(2) & $\bm{0.319} \pm 0.002$ & $0.787 \pm 0.008$ & $\bm{0.618} \pm 0.010$ & $\bm{0.616} \pm 0.005$ \\
 PMC(4) - FNN(2) & $0.321 \pm 0.001$ & $0.780 \pm 0.006$ & $0.634 \pm 0.019$ & $\bm{0.616} \pm 0.004$ \\
 \hline
 FNN(3) & $0.335 \pm 0.004$ & $0.792 \pm 0.007$ & $0.752 \pm 0.036$ & $0.622 \pm 0.002$ \\
 PMC(2) - FNN(3) & $0.322 \pm 0.004$ & $\bm{0.778} \pm 0.009$ & $0.657 \pm 0.026$ & $0.619 \pm 0.003$ \\
 PMC(3) - FNN(3) & $\bm{0.319} \pm 0.003$ & $0.786 \pm 0.007$ & $\bm{0.625} \pm 0.013$ & $\bm{0.609} \pm 0.004$ \\
 PMC(4) - FNN(3) & $0.320 \pm 0.002$ & $0.783 \pm 0.002$ & $0.631 \pm 0.013$ & $0.612 \pm 0.006$ \\
  \hline
 FNN(2, 3) & $0.333 \pm 0.001$ & $0.796 \pm 0.009$ & $0.730 \pm 0.013$ & $0.623 \pm 0.001$ \\
 PMC(2) - FNN(2, 3) & $\bm{0.318} \pm 0.002$ & $\bm{0.775} \pm 0.005$ & $0.643 \pm 0.014$ & $0.624 \pm 0.003$ \\
 PMC(3) - FNN(2, 3) & $0.321 \pm 0.002$ & $0.789 \pm 0.008$ & $\bm{0.639} \pm 0.018$ & $0.616 \pm 0.005$ \\
 PMC(4) - FNN(2, 3) & $0.320 \pm 0.001$ & $0.783 \pm 0.007$ & $0.641 \pm 0.013$ & $\bm{0.614} \pm 0.003$ \\
 \hline
 \hline
 HMC(2) & $0.341 \pm 0.009$ & $0.806 \pm 0.016$ & $0.743 \pm 0.016$ & $\bm{0.624} \pm 0.002$ \\
 HMC(3) & $0.331 \pm 0.003$ & $0.799 \pm 0.007$ & $0.749 \pm 0.033$ & $\bm{0.624} \pm 0.001$ \\
 HMC(4) & $\bm{0.328} \pm 0.002$ & $\bm{0.781} \pm 0.009$ & $\bm{0.724} \pm 0.033$ & $\bm{0.624} \pm 0.002$ \\
\end{tabular}
\captionof{table}{MSE of the different models for volatility forecasting on normalized data}
\label{results_normalized_data}
\end{table}

\begin{table}[t]
\centering
\begin{tabular}{c|c|c|c|c} 
 {} & BTC & SPY & GOLD & EURUSD \\
 \hline
 \hline
 GARCH(1, 1) & $1.226 \pm 0.019$ & $\bm{2.855} \pm 0.013$ & $1.350 \pm 0.015$ & $1.049 \pm 0.000$\\
 PMC(2) - GARCH(1, 1) & $\bm{1.193} \pm 0.004$ & $2.893 \pm 0.016$ & $1.212 \pm 0.016$ & $\bm{1.014} \pm 0.018$ \\
 PMC(3) - GARCH(1, 1) & $1.205 \pm 0.010$ & $2.884 \pm 0.014$ & $\bm{1.164} \pm 0.028$ & $1.034 \pm 0.022$ \\
 PMC(4) - GARCH(1, 1) & $1.199 \pm 0.009$ & $2.870 \pm 0.019$ & $1.197 \pm 0.045$ & $1.049 \pm 0.008$ \\
 \hline
 FNN(2) & $1.201 \pm 0.002$ & $\bm{2.843} \pm 0.009$ & $1.310 \pm 0.016$ & $1.098 \pm 0.001$ \\
 PMC(2) - FNN(2) & $1.198 \pm 0.005$ & $2.896 \pm 0.025$ & $1.188 \pm 0.020$ & $1.087 \pm 0.016$ \\
 PMC(3) - FNN(2) & $\bm{1.193} \pm 0.003$ & $2.881 \pm 0.023$ & $\bm{1.138} \pm 0.016$ & $1.070 \pm 0.022$ \\
 PMC(4) - FNN(2) & $\bm{1.193} \pm 0.003$ & $2.872 \pm 0.017$ & $1.156 \pm 0.019$ & $\bm{1.063} \pm 0.004$ \\
 \hline
 FNN(3) & $1.202 \pm 0.004$ & $2.855 \pm 0.006$ & $1.296 \pm 0.016$ & $1.106 \pm 0.009$ \\
 PMC(2) - FNN(3) & $1.196 \pm 0.005$ & $2.874 \pm 0.013$ & $1.184 \pm 0.033$ & $1.097 \pm 0.008$ \\
 PMC(3) - FNN(3) & $1.195 \pm 0.002$ & $\bm{2.851} \pm 0.014$ & $\bm{1.141} \pm 0.019$ & $1.058 \pm 0.016$ \\
 PMC(4) - FNN(3) & $\bm{1.193} \pm 0.002$ & $2.874 \pm 0.019$ & $1.158 \pm 0.014$ & $\bm{1.041} \pm 0.038$ \\
  \hline
 FNN(2, 3) & $1.211 \pm 0.010$ & $\bm{2.858} \pm 0.022$ & $1.321 \pm 0.022$ & $1.112 \pm 0.008$ \\
 PMC(2) - FNN(2, 3) & $\bm{1.193} \pm 0.005$ & $2.903 \pm 0.038$ & $1.177 \pm 0.018$ & $1.186 \pm 0.124$ \\
 PMC(3) - FNN(2, 3) & $\bm{1.193} \pm 0.002$ & $2.888 \pm 0.034$ & $1.161 \pm 0.019$ & $1.072 \pm 0.007$ \\
 PMC(4) - FNN(2, 3) & $1.195 \pm 0.002$ & $2.874 \pm 0.014$ & $\bm{1.155} \pm 0.016$ & $\bm{1.058} \pm 0.017$ \\
 \hline
 \hline
 HMC(2) & $1.206 \pm 0.005$ & $\bm{2.842} \pm 0.016$ & $1.335 \pm 0.029$ & $1.100 \pm 0.003$ \\
 HMC(3) & $\bm{1.200} \pm 0.003$ & $2.854 \pm 0.007$ & $1.285 \pm 0.021$ & $\bm{1.106} \pm 0.004$ \\
 HMC(4) & $1.201 \pm 0.004$ & $\bm{2.842} \pm 0.020$ & $\bm{1.283} \pm 0.040$ & $1.108 \pm 0.003$ \\
\end{tabular}
\captionof{table}{MSE of the models for volatility forecasting (non-normalized data)}
\label{results_data}
\end{table}

\section{Discussion}

\paragraph{}
As one may observe, except for a few exceptions, extending a model with the PMC procedure allows for improved performance, as theoretically expected. However, the best performances are not necessarily associated with the highest value of $N$. This can be explained, in particular, by the overfitting to which we may be subjected during training.

\paragraph{}
The advantage of models based on PMC is always observable on normalized data, as these are raw training data. Regarding non-normalized data, the situation is more nuanced, as the logarithmic transformation can alter the hierarchy of models.

\begin{figure}[t]
    \centering
    \includegraphics[scale=0.5]{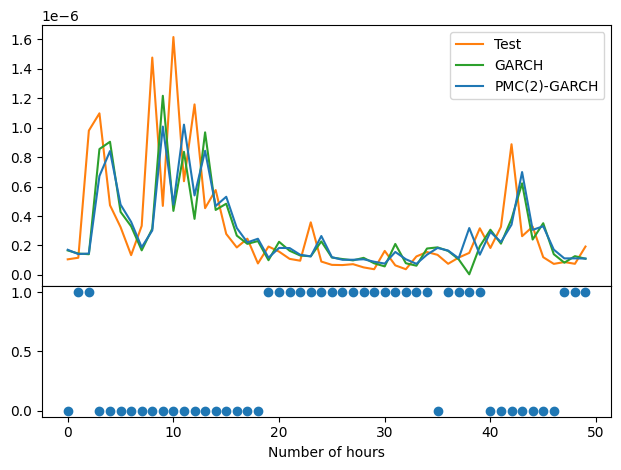}
    \caption{Model predictions with some PMC states}
    \label{pmc_states}
\end{figure}

\paragraph{}
In the context of the results obtained, it is interesting to delve further into the contribution of different states. To illustrate our point, we focus on the comparison between GARCH(1, 1) and PMC(2)-GARCH(1, 1) applied to the BTC data. First, we compare the parameter values. To recall, on the one hand, for the GARCH(1, 1) model, one has the parameters $(\omega, \alpha, \beta)$. On the other hand, with the PMC(2)-GARCH(1, 1), there are two sets $(\omega_0, \alpha_0, \beta_0)$ and $(\omega_1, \alpha_1, \beta_1)$ from two GARCH(1, 1) models aggregated with \ref{pmc_f}. In our case, we have the following values:
\begin{itemize}[itemsep=0em, topsep=0pt, partopsep=0pt]
    \item $(\omega, \alpha, \beta) = (-0.0155, 0.1674, 0.7221)$;
    \item $(\omega_0, \alpha_0, \beta_0) = (0.1730, 0.0161, 0.6508)$ and $(\omega_1, \alpha_1, \beta_1) = (-0.3346, -0.0432,  0.4853)$
\end{itemize}
To illustrate these states, we present some predictions on the test set with the PMC(2)-GARCH(1, 1) in figure \ref{pmc_states}. From the graph, we can infer that state 0 corresponds to low volatility periods, while state 1 indicates high volatility periods. This deduction aligns with the values of the learned parameters.

\section{Conclusion}

\paragraph{}
In this paper, we have developed a new algorithm with the PMC to forecast the next value of a time series. This algorithm is relevant insofar as it allows the introduction of hidden states that make any predictive model non-stationary. Moreover, it does not suffer from the usual features problem of generative models, thanks to a recent methodology described in \cite{azeraf2022deriving, azeraf2023equivalence}. Therefore, this new algorithm allows for the improvement of any forecasting model, and we show that empirically with volatility forecasting.

\paragraph{}
We have applied our new algorithm to volatility forecasting by extending the GARCH(1, 1) model and FNN-based models. As future directions, it might be interesting to extend to other models such as RNNs \cite{rumelhart1985learning, jordan1990attractor}, LSTMs \cite{hochreiter1997long}, or even XGBoost \cite{chen2016xgboost}, and apply them to other forecasting tasks.

\subsection*{About the Author}

Elie Azeraf holds a Ph.D. in Statistics from Institut Polytechnique de Paris, as well as three Master of Science degrees in Financial Engineering, Applied Mathematics, and Data Science. His research interests include finance, time series analysis, and statistical modeling. He is the founder of Azeraf Financial Consulting \href{https://www.azeraf-financial-consulting.com/}{https://www.azeraf-financial-consulting.com/}, a firm dedicated to the design and implementation of portfolio construction strategies tailored to the specific objectives and risk profiles of individual and professional investors.



\bibliographystyle{unsrt}  
\bibliography{references}

\end{document}